# Fighter flight trajectory prediction based on spatio-temporal graph attention network

Yao Sun [a], Tengyu Jing [b], Jiapeng Wang [b], Wei Wang [b,*]

[a] *School of Aeronautical Engineering, Air Force Engineering University, Xi'an, China*
[b] *School of Information and Communication Engineering, Xidian University, Xi'an, China*
[*] *Corresponding author: wwang@mail.xidian.edu.cn*

**Abstract:** Quickly and accurately predicting the flight trajectory of a blue army fighter in close-range air combat helps a red army fighter gain a dominant situation, which is the winning factor in later air combat. However, due to the high speed and even hypersonic capabilities of advanced fighters, the diversity of tactical maneuvers, and the instantaneous nature of situational transitions，it is difficult to meet the requirements of practical combat applications in terms of prediction accuracy. To improve prediction accuracy, this paper proposes a spatio-temporal graph attention network (ST-GAT) using encoding and decoding structures to predict the flight trajectory. The encoder adopts a parallel structure of Transformer and GAT branches embedded with the multi-head self-attention mechanism in each front end. The Transformer branch network is used to extract the temporal characteristics of historical trajectories and capture the impact of the fighter's historical state on future trajectories, while the GAT branch network is used to extract spatial features in historical trajectories and capture potential spatial correlations between fighters. Then we concatenate the outputs of the two branches into a new feature vector and input it into a decoder composed of a fully connected network to predict the future position coordinates of the blue army fighter. The computer simulation results show that the proposed network significantly improves the prediction accuracy of flight trajectories compared to the enhanced CNN-LSTM network (ECNN-LSTM), with improvements of 47% and 34% in both ADE and FDE indicators, providing strong support for subsequent autonomous combat missions.

## 1. Introduction

At present, the development of the air force of each military power gradually presents the trend of integration of mechanization, informatization, and intelligence. Intelligent air combat based on informatization can directly empower and enhance the combat power of the air force, then promote the rapid evolution and development of the combat mode and technical form of air combat, and continuously derive new tactical systems[1]. Due to the continuous improvement of high-speed breakthrough defense capability and stealth performance of fighters, close-range combat is still one of the inevitable forms of current and future air combat. In the close-range intelligent air combat system composed of core technologies such as situational awareness, identification prediction, situational assessment, and maneuver decision-making, the prediction of fighter flight trajectory plays a very crucial role, and its prediction results can provide strong support for the follow-up autonomous combat mission of unmanned aerial vehicles and the auxiliary combat mission of manned aerial vehicles, which can help the red side to prevent risks in the fight, take the advantage of the attack, and then achieve the final victory.

The trajectory prediction problem refers to the use of historical trajectory information and existing knowledge to predict the future position of a target. Currently, there are more applications and research in intelligent traffic such as civil aviation and vehicle autopilot fields, where trajectory prediction is usually used for the research of techniques such as traffic flow estimation and obstacle avoidance for long periods in the future. Early trajectory prediction methods are based on statistical methods and machine learning methods using feature engineering, both of which achieve the prediction of future trajectories by using the features and statistical laws in a large amount of historical trajectory data. However, these methods are limited in information extraction ability and expression ability of features, resulting in poor generalization performance and low prediction accuracy.

With the increasing maturity of deep learning algorithms, significant progress has been made in trajectory prediction, leveraging techniques such as CNN, LSTM, attention mechanisms, and graph networks. Recurrent neural networks represented by RNN and LSTM can learn well for data with temporal characteristics. For example, to ensure the safe and orderly completion of civil aviation flight tasks, Z.Shi et al[2] proposed a constrained long short-term memory network for flight trajectory prediction, which combines the dynamic characteristics of aircraft (climb, cruise, descent) with the LSTM network. Alahi et al. [3] introduced Social LSTM, a model based on social long short-term memory networks. Deo et al. [4] further enhanced this approach by incorporating a convolutional layer to optimize the original social pooling layer, resulting in the network outputting multiple possible future trajectories. Due to the high spatio-temporal correlation of trajectories, Nikhil et al. [5] argued that such data could be more effectively handled using CNNs. They applied a convolutional layer to the sequence structure to capture the spatial correlation in trajectories. As a result, CNNs have demonstrated remarkable performance in trajectory prediction tasks, outperforming RNNs in the time series domain [6]. In addition, variational self-encoder models and Generative Adversarial Networks (GANs) can also be used for trajectory prediction. Such as Walker et al [7] proposed a framework for unsupervised learning in 2014. Gupta et al [8] proposed a Social-GAN network that uses a new aggregation mechanism to aggregate information between people to predict reasonable pedestrian trajectories. Then Li et al [9] proposed a probabilistic trajectory prediction method based on a conditional generative neural system, which firstly inputs historical trajectories and environmental information as conditions into a neural network, and then uses a generative model to generate future trajectories. Subsequently, Kipf et al [10] introduced the concept of CNN into graph neural networks for the first time and proposed a Graph Convolutional Network (GCN), which allows for the efficient extraction of node and edge features. Huang et al [11] proposed a spatio-temporal graph attentional network based on the sequence-to-sequence structure for predicting future pedestrian trajectories.

Chandra et al [12] used a two-layer GNN-LSTM structure to solve the trajectory prediction problem. Huang Zijie [13] proposed a Social-Spatial-Temporal Graph Convolutional Neural Network (Social-STGCNN) to complete the trajectory prediction by focusing on the interaction between pedestrians on the road for modeling. Wang Tianbao [14] proposed the TP-GCN algorithm to address the problem of the difficulty of efficiently constructing pedestrian interactions for the task of pedestrian trajectory prediction. C. Liu [15] proposed a new method, AVGCN, for trajectory prediction based on human attention and improved using variational trajectory prediction with full consideration of the stochastic nature of pedestrian trajectories. The results show that this method has the best performance on several trajectory prediction indicators. H. Jeon [16] used an edge-enhanced graph convolutional neural network to construct an interactive embedding network between vehicles and proposed the first fully scalable vehicle trajectory prediction network SCALE-Net, which can guarantee good prediction performance and consistent calculation load regardless of the number of surrounding vehicles. Y-Fang et al [17] take into account the influence of the surrounding pedestrian movement and use a high-order GCN network to model the interaction between pedestrians, which takes into account both the neighbors of the target pedestrians as well as the neighbors of their neighbors, which in turn can avoid the collision of the prediction results with others.

With the emergence of new technologies such as deep learning, reinforcement learning, and the rapid application of trajectory prediction in the civilian field, air combat flight trajectory prediction methods have also made great progress. Recently, Zhi fei et al. [18] proposed a hybrid algorithm called AERTros-Volterra and introduced an ensemble learning scheme into air combat flight trajectory prediction. The ensemble prediction model is adaptively updated based on the model's performance on real-time samples and the recognition results of target maneuver segmentation points. In addition, multiple achievements have been made in jointly solving trajectory prediction problems with maneuver decision-making, situational assessment, and other related issues: Z. Wei et al [19] designed a TSO-GRU-Ada prediction model, which divides the trajectory prediction model into tactical maneuver prediction as well as trajectory point prediction, and combines the triangular search optimization of the Adaboost and the GRU network. Xie Lei et al [20] proposed a decision-making method that combines the dynamic relational weighting algorithm with the movement time strategy and added the Ada-LSTM trajectory prediction algorithm to the maneuver decision. Although deep learning can effectively extract potential features from Euclidean spatial data, it is still powerless in the face of extensive non-Euclidean data in practical application scenarios, because the data of these application scenarios are generated in non-Euclidean space. Such as non-Euclidean spatial data that is common in air combat environments: graph data, so it is necessary to introduce graph network methods for such scenarios. Y. Sun et al. [21] proposed a prediction algorithm based on attention mechanism and graph convolutional networks, which pioneered the addition of graph convolutional networks to reflect the positional relationships between different fighter jets, achieving trajectory prediction during multi-fighter confrontation.

This paper draws on the successful experience of Transformer networks in the field of time series prediction [22], and is inspired by Jiansen Zhao's [23] use of GAT networks to extract spatial features of ship trajectories for accurate prediction. Combining the mission characteristics and requirements of close-range fighter combat, this paper proposes a spatio-temporal graph attention network to achieve flight trajectory prediction.

## 2. Modelling framework

The overall network structure of this paper is shown in Fig. 1, the model as a whole is divided into two parts: encoder and decoder. On the encoder side, two fully connected layers FC1, FC2 are used to transform and adjust the raw data in the trajectory respectively. The Transformer module is used to extract features from the historical flight trajectory $traj = \{traj_1, traj_2, traj_3, \cdots, traj_n\}$ of the fighter in the time dimension, in which the purely self-attention network is used to predict the mutated trajectories in the complex and changing trajectory scenarios. At the same time, the GAT module is used to extract features of the spatial dimension of the flight trajectory of the fighter, the temporal and spatial features are spliced and sent to the fully connected network FC3 for decoding to obtain the coordinate values of $x$, $y$, $z$ and at the next moment. The sliding window method is utilized as shown by the long black arrow in the figure. The new output points are used as known trajectory points and spliced with the historical trajectory of the previous moment as inputs for the next round of prediction, until the prediction of all target points is completed.

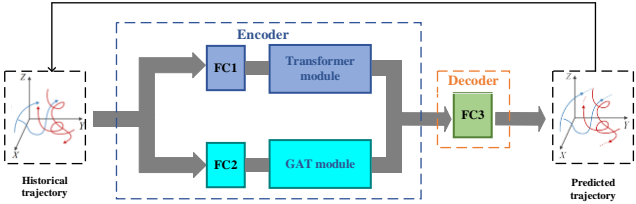

Fig. 1 Spatio-temporal graph attention network model

## 2.1 Transformer module

Referring to the article [24], this paper uses the encoder part of the Transformer network structure for temporal feature extraction of trajectory information. The Transformer module outputs a feature vector of dimension $l \times n \times 24$, where $l$ denotes the length of the historical trajectory, and $n$ denotes the number of fighters. The specific structural parameters in the Transformer module are shown in the following table, PE denotes position encoding, MHA denotes multi-head attention, FNN denotes feed-forward network, LN denotes normalization layer.

Table 1 Transformer module structure parameters

| Network layer | heads | Input size | Drop out_rate | Output size |
|---|---|---|---|---|
| PE | — | $l \times n \times 24$ | 0 | $l \times n \times 24$ |
| MHA | 4 | $l \times n \times 24$ | 0.1 | $l \times n \times 24$ |
| FNN | — | $l \times n \times 24$ | 0.1 | $l \times n \times 24$ |
| LN | — | $l \times n \times 24$ | 0 | $l \times n \times 24$ |

## 2.2 GAT module

In air combat scenarios, traditional methods cannot effectively share the flight characteristics of fighters in many to many air combat environments. Therefore, this paper uses the GAT module to aggregate the spatial correlation of historical trajectories using the relationship between vertices and edges in the graph structure.

Each fighter is characterized by $h_i$, and the dimension of $h_i$ is $8 \times 3$. It is fed into the graph network to calculate the attention weight of the current node concerning the other nodes with the following formula:

$$e_{ij}^t = \alpha(Wh_i, Wh_j) \quad (1)$$

In Equation (1), the features of different nodes are dimensionally expanded using a weight matrix $8 \times 24$ to enhance their feature expression. Subsequently, the dimension-expanded features are spliced and transformed into a constant for representing the attention coefficient by $\alpha$ matrix operation. To better characterize the relationship between different nodes, it can be normalized using the $softmax$ function:

$$\alpha_{ij} = softmax_i(e_{ij}) \quad (2)$$

Equation (2) is modified to obtain a new formula for calculating the attention weights:

$$\alpha_{ij} = \frac{\exp(Leaky\operatorname{Re}LU(a^T[Wh_i \| Wh_j]))}{\sum_{k \in N} \exp(Leaky\operatorname{Re}LU(a^T[Wh_i \| Wh_{jk}]))} \quad (3)$$

Where $\|$ denotes the splicing operation and $N$ denotes the node adjacent to node $i$.

The formula for the feature vector after multi-attention fusion is:

$$h_{ij}^{''} = \sigma(\frac{1}{K}\sum_{i=1}^{K} h_i^{'}) \quad (4)$$

where $h_i^{''}$ and $h_i^{'}$ denote the features after fusion of the multi-head attention mechanism and the features of the single attention mechanism, respectively, and $\sigma$ is the Sigmoid activation function. The dimension of the final GAT module output feature vector is $8 \times n \times 24$, and $n$ is the number of fighters.

## 3. Results

### 3.1 Description of the dataset

In this paper, thirty typical air battles are selected on the DCS World platform, including eight "4vs4", twelve "2vs2" and ten "1vs1" battles. The sampling frequency is 2 Hz and each air battle has 300 to 3,800 data items, each of which records the flight data such as the altitude, speed, longitude, latitude, roll angle, yaw angle, pitch angle, etc. After the obtained data were low-pass filtered to filter out the noise interference they are divided into a training set and a test set in the ratio of 4:1.

### 3.2 Development environment

This paper uses Python language and PyTorch framework for model construction and simulation experiments, and adpot a dual-card crossfire 1080ti graphics card host as the experimental hardware environment. The integrated development environment is PyCharm, and the deep learning framework is Pytorch 1.3.1.

### 3.3 Evaluation indicators

(1) ADE(Average Displacement Error): Used to measure the average error of the trajectory prediction model throughout the prediction process, calculated as follows:

$$ADE = \frac{1}{n}\sum_{i=1}^{n}\frac{1}{t_{pred}}\sum_{t=t_{obs}+1}^{t_{obs}+t_{pred}}\sqrt{\begin{array}{l}(x_i^t - \hat{x}_i^t)^2 + \\ (y_i^t - \hat{y}_i^t)^2 + \\ (z_i^t - \hat{z}_i^t)^2\end{array}} \quad (5)$$

where the dataset contains a total of $n$ moments; $t_{obs}$ represents the trajectory data containing $obs$ historical moments during the prediction process; $x_i^t$, $y_i^t$ and $z_i^t$ represent the Cartesian coordinate values (true values) of fighter $i$ at moment $t$, while $\hat{x}_i^t$, $\hat{y}_i^t$ and $\hat{z}_i^t$ represent the Cartesian coordinate values (predicted values) of fighter $i$ predicted by the network at moment $t$, respectively.

(2) FDE(Final Displacement Error): The Euclidean distance between the endpoint of the predicted trajectory and the endpoint of the true trajectory. Its calculation formula is as follows:

$$FDE = \sqrt{\begin{array}{l}(x_i^{t_{obs}+t_{pred}} - \hat{x}_i^{t_{obs}+t_{pred}})^2 + \\ (y_i^{t_{obs}+t_{pred}} - \hat{y}_i^{t_{obs}+t_{pred}})^2 + \\ (z_i^{t_{obs}+t_{pred}} - \hat{z}_i^{t_{obs}+t_{pred}})^2\end{array}} \quad (6)$$

where $x_i^{t_{obs}+t_{pred}}$, $y_i^{t_{obs}+t_{pred}}$ and $z_i^{t_{obs}+t_{pred}}$ represent the Cartesian coordinate values of fighter $i$ at the last moment (true values), and $\hat{x}_i^{t_{obs}+t_{pred}}$, $\hat{y}_i^{t_{obs}+t_{pred}}$ and $\hat{z}_i^{t_{obs}+t_{pred}}$ represent the Cartesian coordinate values of fighter $i$ predicted by the network at the last moment (predicted values), respectively.

### 3.4 Comparative experimental analysis

3.4.1. Comparison of spatio-temporal graph attention network and the enhanced CNN-LSTM network in different scenarios

To make a more comprehensive comparison, two different engagement scenarios are selected to be analyzed in this paper, namely: "2vs2", "4vs4". The historical trajectories of different environments are used as inputs and fed into the two networks respectively to obtain the output trajectories of the two sets of networks, to measure the advantages and disadvantages of this paper's method and the enhanced CNN-LSTM network[25]. In the following two sets of images, the red part represents the trajectory of the past 8 moments, the blue part represents the real future trajectory, and the green dotted line represents the future trajectory predicted by the model based on the historical trajectory. Figure (a) both show the predicted trajectory graphs of the spatio-temporal graph attention network. Figure (b) both show the predicted trajectories of the enhanced CNN-LSTM network. When reading the graphs the x-axis needs to add 3590 km, the z-axis needs to add 4100 km, in Fig. 4 the y-axis needs to add 3220 km, and in Fig. 5 the y-axis needs to add 3200 km.

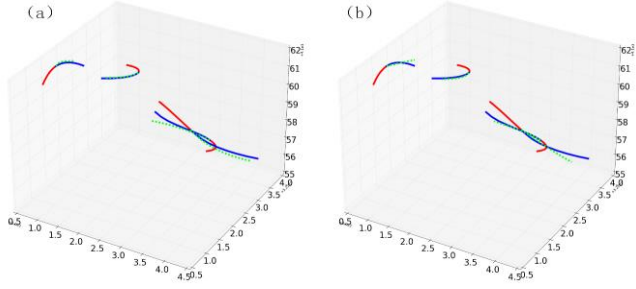

Fig. 2 Comparison of the prediction results of different networks in "2vs2" combat scenarios

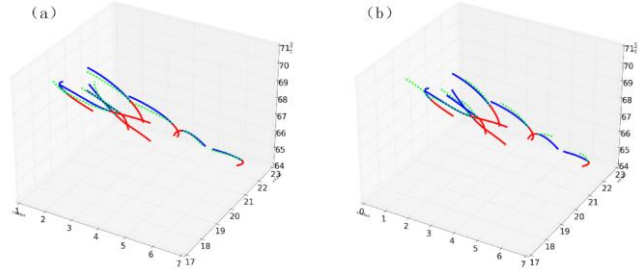

Fig. 3 Comparison of the prediction results of different networks in "4vs4" combat scenarios

As shown in Fig. 2, in the "2vs2" engagement scenario, both the enhanced CNN-LSTM network and the spatio-temporal graph attention network can correctly judge the future flight direction of the fighters from the information of the historical trajectories, and predict the flight trajectories of the fighters in a certain moment. However, on the right side of Fig. 2(a) and Fig. 2(b), the prediction error of the spatio-temporal graph attention network is smaller than that of the enhanced CNN-LSTM network when there exists a more intense confrontation.

As shown in Fig. 3, in the "4vs4" engagement scenario, the enhanced CNN-LSTM network has a slightly poorer ability to capture the spatial relationship between the fighters, which leads to a lower accuracy of the method in predicting complex flight trajectories. The spatio-temporal graph attention network extracts spatio-temporal features through Transformer and GAT, and utilizes the relationship between points and edges in the graph data structure to simulate the spatial positional change relationship of fighters. This network can better extract the interaction features in complex historical trajectories and make the predicted trajectories closer to the real trajectories.

From the perspective of quantitative analysis, the ADE, FDE (in kilometers) of the output results of the two networks in the above air combat environments are recorded as shown in Table2, 3.

Table 2 "2vs2" air combat environments

| Module | ADE | FDE |
|---|---|---|
| ECNN-LSTM | 0.185 | 0.197 |
| ST-GAT | 0.098 | 0.124 |

Table 3 "4vs4" air combat environments

| Module | ADE | FDE |
|---|---|---|
| ECNN-LSTM | 0.465 | 0.563 |
| ST-GAT | 0.239 | 0.313 |

After observing the predicted trajectory graphs and analyzing the evaluation indicators, it can be analyzed that the spatio-temporal graph attention network in different air combat environments has higher accuracy compared to the enhanced CNN-LSTM network using the attention mechanism, and the predicted trajectories are closer to the real trajectories.

3.4.2. Effect of sampling frequency on the method of this paper

The sampling frequency used in this paper to obtain trajectories from air battles is 2Hz. The distance difference between the interval points is relatively large. To verify the feasibility of the algorithm in this paper, this section uses the same air battle scenario with the sampling frequency of 2Hz and 50Hz to conduct comparative experiments. When the sampling frequency is 2Hz, the historical 8 trajectory points are used to predict the next trajectory point. At a sampling frequency of 50Hz, a history of 200 trajectory points is used to predict the next trajectory point. Both use the first 4 seconds of trajectory information.

When the sampling frequency is 2Hz, the error of predicting the next trajectory point is 0.009 km by using 8 trajectory points in history; when the sampling frequency is 50Hz, the error of predicting the next trajectory point is 0.0078 km by using 200 trajectory points in history. It is verified that increasing the sampling frequency to 50Hz helps the method in this paper to achieve better prediction results. However, with the sampling frequency increasing, the number of history points used at the same time increases dramatically, the computation volume increases, and the accuracy improvement is small.

*3.5 Ablation experiment*

To better reflect the effectiveness of this paper's method, this paper adopts a Transformer branch network, GAT branch network, and spatio-temporal graph attention network to carry out experiments under the same experimental conditions. To more intuitively reflect the changing trend of trajectory, the trajectory points of 8 moments in history are taken to predict the trajectory points of 8 moments in the future. Experimental results are shown as follows.

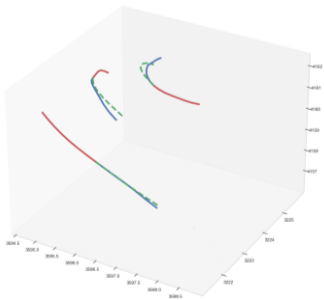

Fig. 4 Spatio-temporal graph attention network

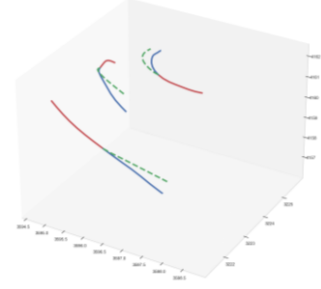

Fig. 5 GAT branch network

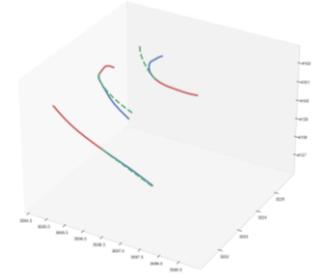

Fig. 6 Transformer branch network

From the figures, it can be seen that the Transformer branch network is well for temporal information, but the prediction accuracy decreases when there is a sudden change in the flight attitude of the fighter. The GAT branch network can predict the change in the flight trajectory of the fighter in time due to its strong spatial extraction ability. The method in this paper combines the advantages of the two, fuses the temporal and spatial features, and produces better prediction results.

From the perspective of quantitative analysis, this paper records the ADE, FDE of predicting the trajectory points of one moment in the future from the trajectory points of eight moments in history under different networks under the ablation experiments. The values of ADE and FDE are the same when only one point is predicted. The ADE is 0.009 km in the spatio-temporal graph attention network, 0.021 km in the Transformer branch network, and 0.094 km in the GAT branch network. In summary, the algorithm in this paper performs best.

## 4. Discussion

In this paper, the graph structure is introduced into the prediction of air combat trajectory, and the GAT network is combined with the Transformer network for constructing the spatio-temporal graph attention network. Through comparison experiments with extensive groups of networks, it is verified that the spatio-temporal graph attention network has a better prediction effect. The current spatio-temporal graph attention network can extract features in both spatial and temporal dimensions, but whether the network structure can be further optimized to analyze the pilot's flight intention by using multimodal fusion so that it can be better adapted to the characteristics of the air combat flight trajectory prediction mission, is worth exploring in depth.

**CRediT authorship contribution statement**

Yao Sun: Methodology, Validation, Investigation, Formal analysis, Writing-original draft. Tengyu Jing and Jiapeng Wang: Conceptualization. Wei Wang: Writing-review & editing.

**Declaration of Competing Interest**

The authors declare that they have no known competing financial interests or personal relationships that could have appeared to influence the work reported in this paper.

**Acknowledgments**

This work was supported by FUND (Grant No. XXX)